\documentclass[runningheads]{llncs}
\usepackage{multirow}
\usepackage{booktabs}
\usepackage{adjustbox}
\usepackage[T1]{fontenc}
\usepackage{graphicx}
\usepackage{caption}
\usepackage{amsmath,amssymb,amsfonts}
\usepackage{subcaption}
\usepackage{graphicx}
\usepackage{textcomp}
\usepackage{xcolor}
\usepackage{bm}

\usepackage{amssymb}
  \usepackage{amsthm}
\begin{document}
\title{CFMD: Dynamic Cross-layer Feature Fusion for Salient Object Detection}
%
%
\titlerunning{Mamba DyUpsampling Enhancement Significant Target Detection in CFPN}

\author{Jin Lian\inst{1}\orcidID{0009-0002-1048-5307} \and
Zhongyu Wan\inst{1}\orcidID{009-0007-6988-6224} \and
MIngGao\inst{1}\orcidID{009-0004-6662-9992} \and
JunFeng Chen\inst{2}\orcidID{009-0008-9694-5725}
}
\authorrunning{Jin Lian et al.}
\institute{Office of Network and Digital Construction, JiangHan University, Wuhan, Hubei, China \and
School of Artificial Intelligence,JiangHan University, Wuhan, Hubei, China
\email{\{wanzhongyu\}@jhun.edu.cn}}
%
\maketitle     \text{These authors contributed equally to this work}  
\begin{abstract}
Cross-layer feature pyramid networks (CFPNs) have achieved notable progress in multi-scale feature fusion and boundary detail preservation for salient object detection. However, traditional CFPNs still suffer from two core limitations: (1) a computational bottleneck caused by complex feature weighting operations, and (2) degraded boundary accuracy due to feature blurring in the upsampling process. To address these challenges, we propose CFMD, a novel cross-layer feature pyramid network that introduces two key innovations.
First, we design a context-aware feature aggregation module (CFLMA), which incorporates the state-of-the-art Mamba architecture to construct a dynamic weight distribution mechanism. This module adaptively adjusts feature importance based on image context, significantly improving both representation efficiency and generalization.
Second, we introduce an adaptive dynamic upsampling unit (CFLMD) that preserves spatial details during resolution recovery. By adjusting the upsampling range dynamically and initializing with a bilinear strategy, the module effectively reduces feature overlap and maintains fine-grained boundary structures.
Extensive experiments on three standard benchmarks using three mainstream backbone networks demonstrate that CFMD achieves substantial improvements in pixel-level accuracy and boundary segmentation quality, especially in complex scenes. The results validate the effectiveness of CFMD in jointly enhancing computational efficiency and segmentation performance, highlighting its strong potential in salient object detection tasks.
\end{abstract}
%
%
%


\section{Introduction}
Feature Pyramid Networks (FPNs) have become a cornerstone in multi-scale feature extraction since their introduction in 2016. By aggregating semantic features across layers through top-down and lateral connections, FPNs effectively capture both low-level details and high-level semantics. Subsequent works, such as Pyramid-based Interaction Networks (PAFAN)~\cite{choudhary2025syntheticfeatureaugmentationimproves,corley2025deepeegsuperresolutionupsampling}, enhance cross-scale correlations by introducing explicit interaction modules, achieving improvements in tasks such as edge-aware object detection.

Despite their success, conventional FPN-based methods face two key challenges: (1) limited interaction between deep semantic features and shallow spatial details due to static fusion structures, which hinders the accurate modeling of object boundaries; and (2) reliance on manually designed fusion strategies or fixed architecture search~\cite{ghiasi2019nasfpnlearningscalablefeature,cui2025detectiongeographiclocalizationnatural}, which limits adaptability to diverse scenes.

Cross-layer Feature Pyramid Networks (CFPNs)~\cite{jafari2025mambalrpexplainingselectivestate} attempt to mitigate these issues by introducing explicit connections and attention-based aggregation. However, they still suffer from two limitations: the flattening of 3D feature maps disrupts spatial structure~\cite{jiakun2024bafpnbidirectionalalignment}, reducing attention ~\cite{shen2025long}accuracy, and fixed upsampling strategies cause spatial misalignment and boundary blur~\cite{kamble2025enhancedmulticlassclassificationgastrointestinal}.

To address these limitations, we propose \textbf{CFMD}, a novel Cross-layer Feature Modulation and Distribution framework, which introduces two key innovations:
We embed the Mamba architecture at the backbone's tail to build a context-aware dynamic weighting mechanism. Global features are extracted via average pooling and modeled with structured state-space models (SSMs)~\cite{shen2023pbsl,shen2023triplet}. Mamba enables long-range dependency modeling and efficient context aggregation via an adaptive scan strategy~\cite{shen2025long}. The resulting attention weights are used to recalibrate multi-scale features, improving semantic fusion.
To reduce spatial distortion during resolution recovery, we design an adaptive upsampling module based on learned offsets. Each scale-level feature map $F^n$ is modulated by content-aware offsets $O_n$:
\begin{equation}
  O_n = 0.5 \cdot \mathrm{sigmoid}(W_1 F_n) \cdot W_2 F_n, \quad 
  F_{\mathrm{up}}^n = \mathrm{grid\_sample}(F_n, G_n + O_n),
\end{equation}
where $W_1$, $W_2$ are learnable weights and $G_n$ is a bilinear grid. This approach enhances detail preservation and maintains spatial consistency across scales.

By integrating CFLMA and CFLMD, our CFMD framework forms a two-stage pipeline (see Fig.~\ref{fig2}) that jointly optimizes efficiency and segmentation precision. Our method is architecture-agnostic and compatible with recent saliency frameworks~\cite{shen2024imagpose,shen2024imagdressing}.
The main contributions of this paper are summarized as follows:

\begin{itemize}
  \item We propose CFMD, a novel framework that introduces Mamba-based modulation and dynamic upsampling for cross-layer feature fusion.
  \item We design CFLMA and CFLMD modules to enhance semantic consistency and spatial fidelity across scales.
  \item Extensive experiments demonstrate that CFMD achieves state-of-the-art performance in salient object detection, particularly in boundary-aware and multi-scale settings.
\end{itemize}


\section{Related Work}
\label{Related work}
\subsection{Cross-layer Feature Pyramids in Salient Object Detection}
Feature Pyramid Networks (FPNs) have become foundational in multi-scale visual representation. Traditional hand-crafted pipelines—such as Haar + AdaBoost or HOG + SVM—were widely used for object detection prior to the deep learning era~\cite{zheng2025xfmambacrossfusionmambamultiview}. However, these approaches lacked robustness and adaptability in complex scenes. The introduction of FPNs enabled hierarchical aggregation of semantic and spatial features, greatly improving detection performance.
Building upon this, PAFPN~\cite{xiao2024globalfeaturepyramidnetwork} incorporated path aggregation to better propagate detailed spatial information from low-level layers. More recently, CFAN~\cite{fu2025llmdetlearningstrongopenvocabulary} introduced dynamic layer weighting to better align boundary and semantic features. These developments significantly enhance salient object detection by capturing finer-grained boundary information and improving robustness to scene variability.

\subsection{State Space Modeling and Mamba}

The Mamba architecture, derived from the State Space Model (SSM), has recently emerged as a powerful alternative to Transformer-based sequence models~\cite{zheng2025xfmambacrossfusionmambamultiview,yang2025mhafyolomultibranchheterogeneousauxiliary}. By replacing token-level attention with selective and structured recurrence, Mamba compresses feature representations for faster and more memory-efficient modeling.
In vision tasks, Vision Mamba processes an image as a 2D sequence of spatial blocks, which are linearly projected, positionally encoded, and passed through bi-directional SSM encoders. This approach achieves accurate, efficient visual understanding with reduced computational overhead~\cite{jafari2025mambalrpexplainingselectivestate}. Inspired by recent diffusion models that progressively guide generation with rich structural priors~\cite{shen2023advancing,shen2024boosting}, Mamba introduces dynamic, context-aware sequence modeling that supports global dependency modeling while remaining efficient—a valuable property for feature fusion in cross-layer architectures.

\subsection{Dynamic Upsampling Strategies}
Traditional fixed upsampling mechanisms often cause boundary blur and spatial misalignment during feature resolution recovery. To overcome this, dynamic upsampling techniques have been proposed. Instead of statically interpolating features, these methods learn spatially adaptive offsets based on content-aware cues, allowing more accurate restoration of spatial details~\cite{kamble2025enhancedmulticlassclassificationgastrointestinal,liu2025efficientpointcloudsupsampling}.
In recent salient object detection frameworks, such strategies enable context-aware adjustment of upsampling regions, improving both local detail recovery and global semantic alignment. Similar to adaptive generation in diffusion-based pipelines~\cite{shen2024boosting}, dynamic upsampling allows for spatial flexibility that is critical in handling variable object scales and shapes.
\section{Method}
\subsection{Overall Architecture}
\begin{figure*}[t]
\begin{center}
\includegraphics[width=0.9\textwidth]{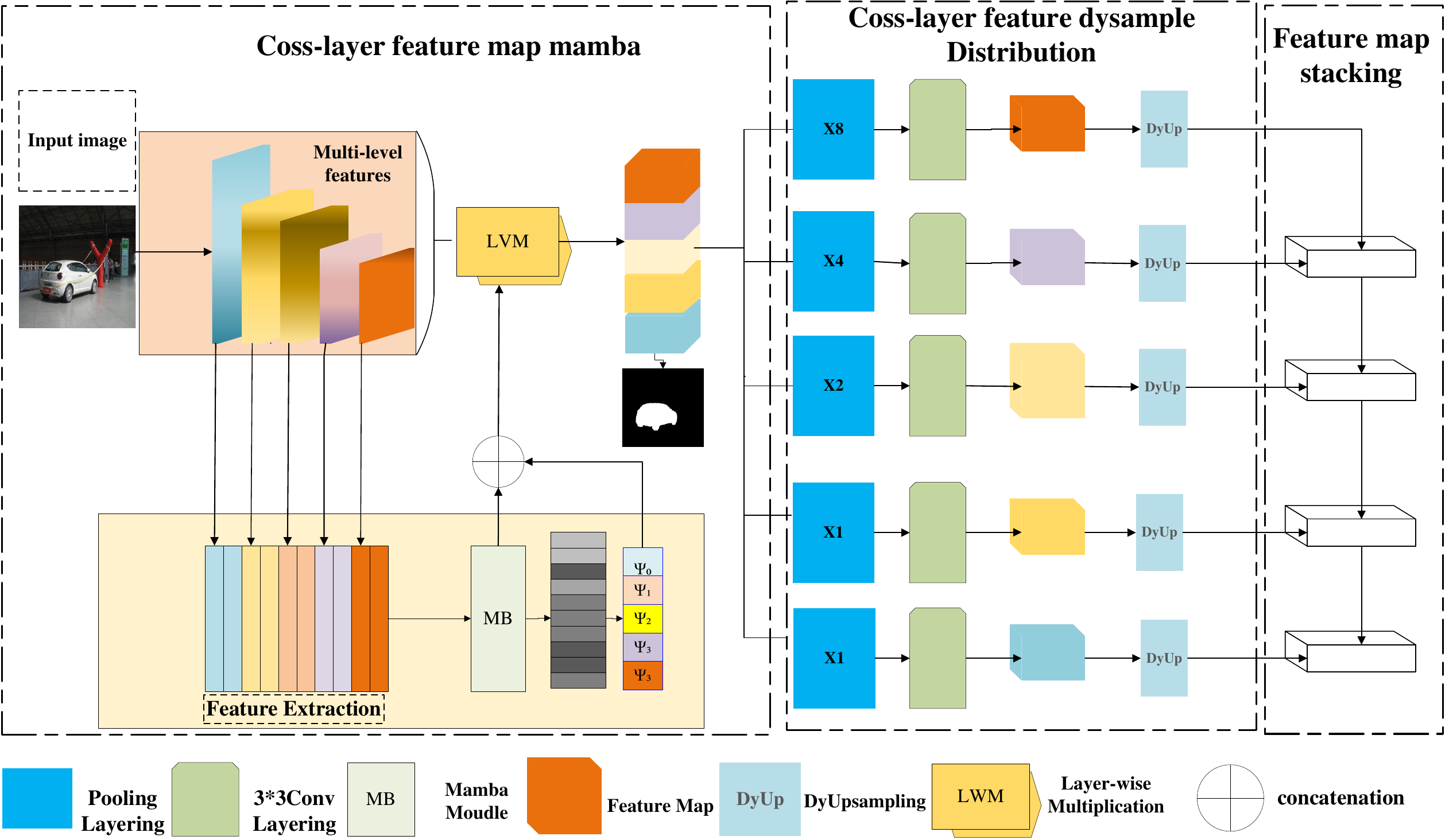}
\vspace{-0.1mm}
\captionsetup{margin=6pt,justification=justified}
\caption{The Cross-layer Feature mamba Aggregation module proposed in this paper introduces a dual dynamic mechanism based on CFPN: the Mamba module replaces the traditional global pooling layer and utilizes the state-space model (SSM) to capture the feature long-range dependency and generates the channel attention mask to realize the dynamic calibration; at the same time, it adopts the nonlinear convolution and the dynamic offset generation technique to construct the dynamic upsampling 
module~\cite{tang2025vpnextrethinkingdense}~\cite{rando2025serpentselectiveresamplingexpressive}~\cite{qiu2025sparsemambapclscribblesupervisedmedicalimage}, which performs fine scanning (step size 1×1) in the high-entropy region (target boundary) and coarse-grained sampling (step size 2×2) in the low-entropy background region. This design significantly improves the adaptability and semantic alignment accuracy of feature fusion, all while maintaining a reassuringly lightweight architecture.} \label{fig2}
\end{center}
\vspace{-3mm}
\end{figure*}
The CFMD architecture shown in  consists of Cross-layer Feature Mamba Aggregation (CLFMA) and Cross-layer Feature Dyupsampling Distribution (CLFDD). The architecture realizes dynamic computation of feature weights and cross-layer fusion through the state space mechanism of the Mamba model see Fig.~\ref{fig2}: firstly, the CLFMA module extracts the long-range dependency relationship between multi-scale features using the state space model and generates an adaptive weight matrix to weight and fuse the feature maps; subsequently, the CLFDD module performs spatial dimensional reconstruction for the CLFMA-processed features through the dynamic upsampling strategy and combines with the state space module's state-space model to generate a dynamic up-sampling distribution. Subsequently, the CLFDD module reconstructs the spatial dimension of the CLFMA-processed features through a dynamic up-sampling strategy and combines it with the noise suppression property of the state-space module to realize the fine filtering of features. Finally, the architecture adopts the top-down feature fusion path similar to CFPN and accomplishes the efficient integration of multi-scale semantic information through the multi-level feature pyramid structure.
\par\begin{figure}[t]
\centering
\includegraphics[scale=0.5,width=0.7\textwidth,clip]{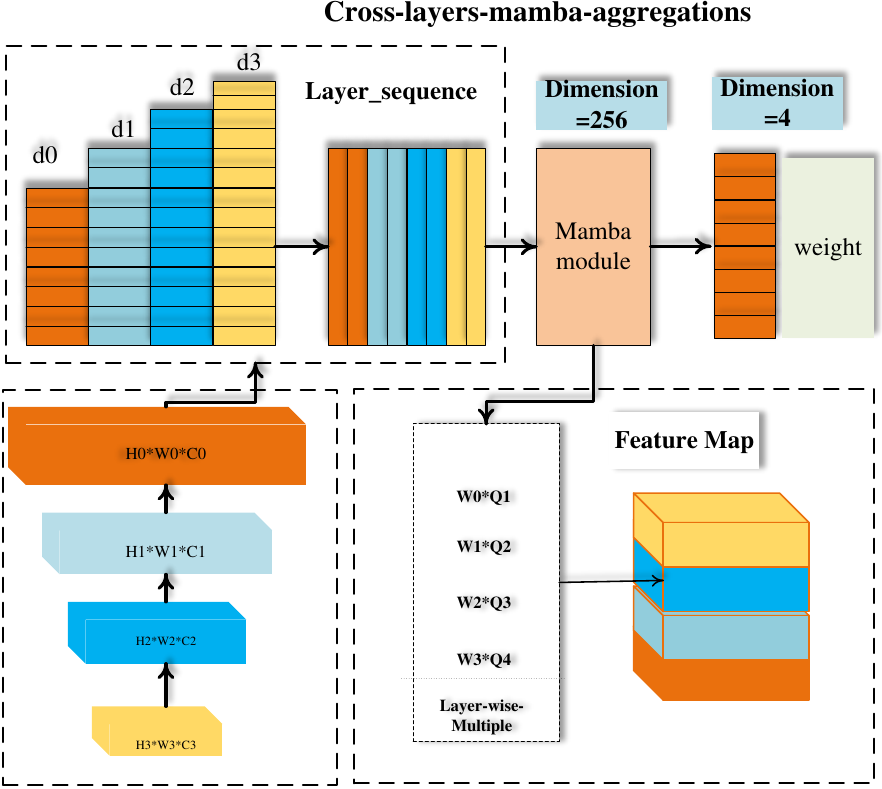}
\captionsetup{margin=6pt,justification=justified}
\vspace{0.1mm}
 \caption{The feature maps are processed through the global pooling layer and scaled to different sizes to produce images of sizes d0, d1, d2, and d4, respectively. Next, these images are feature extracted using a key component in our process-the Mamba module. This module is responsible for extracting the features from the images. The extracted features are then fused with the corresponding weights ($\Psi_0$, $\Psi_1$, $\Psi_2$, $\Psi_3$). The fused feature information is fed into the Output global pooling layer for subsequent processing and output of the final results.}\label{fig3}
\vspace{-2mm}
\end{figure}
\vspace{2mm}

\subsection{Cross-layer Feature mamba Aggregation}
\begin{figure}[t]
\centering
\includegraphics[scale=0.8,width=0.7\textwidth,clip]{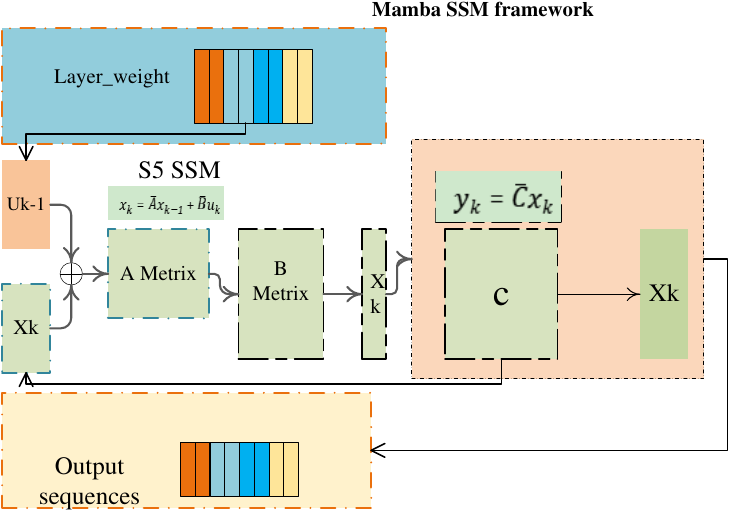}
\vspace{0.1mm}
\captionsetup{margin=6pt,justification=justified}
 \caption{
The structure employs the selective state scanning mechanism of the Mamba module, where the multilevel feature map is first downscaled and converted into a serialized form for input to the module. Subsequently, the core state transfer equation \(x_k = \bar{A}x_{k-1} + \bar{B}u_k\) is used to perform the temporal feature computation, where the dynamically generated parameter matrices \(\bar{A}\) and \(\bar{B}\) are responsible for capturing the long-range spatial dependencies. The computation process sequentially realizes the content-aware feature enhancement through the gating mechanism and utilizes the transmitter module \(y_k = \bar{C}x_k\) to complete the transition from hidden state to observable features. The final processed sequence data is restored to the original spatial dimension \(\mathbb{R}^{B \times 256 \times H \times W}\) by the dimension restoration operation, which is used as an input to the state model for the next moment.}\label{fig4}
\vspace{-2mm}
\end{figure}

Cross-layer Feature mamba Aggregation: Let's delve into the deep learning architecture of the ResNet50 pre-trained model, using it as a case study to understand the data processing flow shown in (see Fig.~\ref{fig3}). 
This flow incorporates several key operations. The input features initially enter the data shape change module, presented as a tensor with B representing the batch size,in\_ch the number of input channels, and H and W the height and width of the feature map. After the feature fusion channel adjustment operation, the number of input channels in\_ch is uniformly adjusted to 256, and the data shape is changed. To meet the subsequent sequence modeling requirements, the spatial features are further flattened into sequence data, as shown in (see  Fig.~\ref{fig4})
\par
\vspace{2mm}
Then, we enter the processing flow of the Mamba module, a key innovation integrated into the core of Cross-layer Feature Aggregation (CFA).The Mamba module adopts a selective state scanning mechanism to capture long-range spatial dependencies through the dynamically generated A/B/C matrix, as shown in Fig.5(see Fig~\ref{fig5}). Its core computation follows the state transfer equation: where is the state of the previous moment, is the current input sequence, and is the dynamically generated parameter matrix. The introduction of gating mechanisms achieves content-aware feature enhancement through the launch module: the hidden state is converted to observable features for the output matrix. The processed sequence data is finally restored to the original spatial structure, and the shape is restored for subsequent cross-layer fusion. The introduction of the Mamba module significantly improves the model's ability to deal with complex scenes (e.g., occlusion, small objects) and, at the same time, maintains a high inference efficiency through its linear computational complexity.
\subsection{Cross-layer Feature dyupsampling Distribution}
\begin{figure}[htbp]
\centering
\includegraphics[scale=0.5,width=0.7\textwidth,clip]{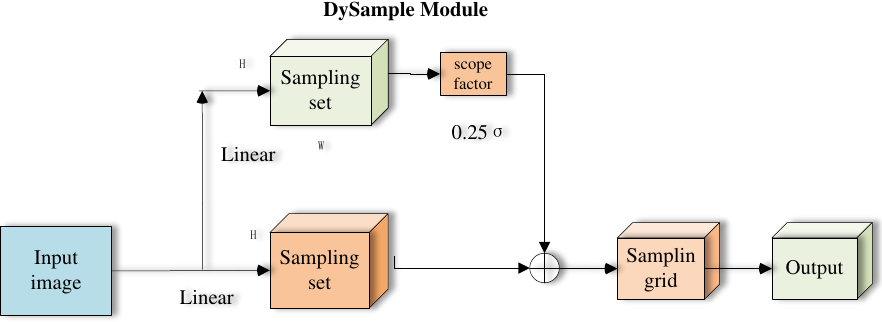}
\vspace{-0.1mm}
\captionsetup{margin=6pt,justification=justified}
 \caption{
To address the up-sampling problem, we have converted it to point sampling and implemented it with PyTorch built-in functions. Our method involves initially generating offsets by linear projection combined with bilinear interpolation. In our improvement, we have introduced the concept of bilinear initialization to enhance accuracy. We have also introduced the static dynamic range factor, where $\sigma$ is set to be a constrained offset in the interval of 0.25. The Sampling set is responsible for converting the information into point sampling units, and the GridSampling is used to aggregate the sampling units to reduce the overlap of the sampling points. We have used feature grouping to generate the sample points independently, thereby enhancing the flexibility of our method. Lastly, we have compared “Linear + Pixel Blending” and “Pixel Blending + Linear” to find the optimal solution and improve the model performance.
 }\label{fig5}
\vspace{-2mm}
\end{figure}

On this basis, a lightweight Dynamic Upsampling Module (DUM) is proposed, whose structure is shown in (see Fig.~\ref{fig5}).
\vspace{2mm}
The feature processing flow of the module starts from the global feature F\_agg generated by the CFA module, and then different scale feature maps are obtained by CFD multi-scale pooling (with r taking the value of 1, 1, 2, 2, 4, 8), and then each feature map is subjected to the DySample dynamic sampling~\cite{zheng2025xfmambacrossfusionmambamultiview} operation respectively to generate the upsampled feature maps, which ultimately constitutes the feature pyramid.
The core design of DUM consists of a nonlinear feature transformation unit and a dynamic offset generation mechanism, which aims to solve the problem of semantic misalignment and detail loss in multi-scale feature fusion. Specifically, given an input feature map \(F_{\text{in}} \in \mathbb{R}^{C \times H \times W}\), the module firstly constructs a non - linear Transformation branch by cascading 1×1 convolution - non - linear activation - 3×3 depth separable convolution, as shown in . Branch), as shown in Fig. 6 (see Fig.~\ref{fig5})
\begin{figure}[htbp]
\centering
\includegraphics[scale=0.5,width=0.7\textwidth,clip]{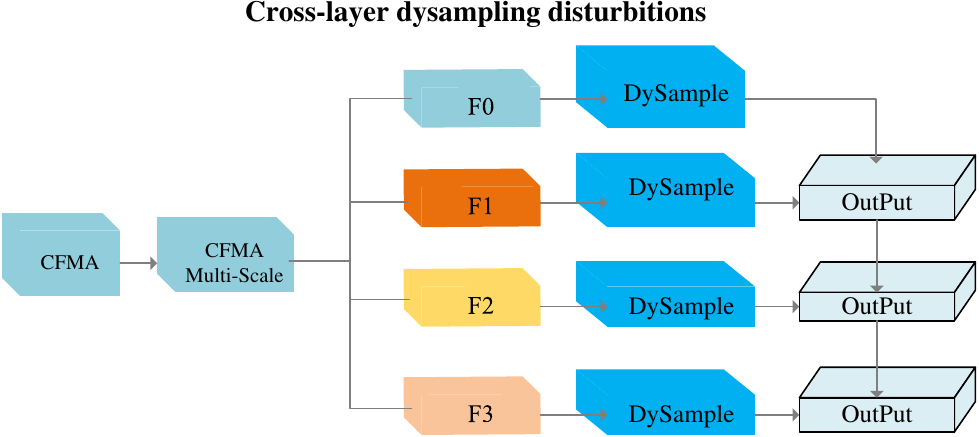}
\captionsetup{margin=6pt,justification=justified}
 \caption{
This figure shows the deep learning feature processing flow: the CFMA module obtains the data to obtain the global feature \(F_{agg}\), and then by the CFMD multi-scale pooling to obtain multiple feature maps. Each feature map uses DySample to generate offsets and dynamically upsampling, and finally fused into a multilevel feature pyramid.}
 \label{fig6}
\vspace{-2mm}
\end{figure}
\vspace{2mm}
, to generate the intermediate features \(F_{\text{mid}} \in \mathbb{R}^{C' \times H \times W}\), with the expression: 
\begin{equation}
F_{\text{mid}} = \text{DWConv}_{3\times3}(\text{GeLU}( \text{Conv}_{1\times1}(F_{\text{in}}))). \end{equation} Where \(\text{DWConv}_{3\times3}\) denotes the depth separable convolution, and \(\text{GeLU}\) is the Gaussian error linear unit activation function, (see Fig 5.~\ref{fig5})
this design enhances the feature discriminative properties by nonlinear mapping while utilizing the depth convolution to reduce the computational complexity.
Subsequently, the Dynamic Offset Branch (DOB) predicts the spatially adaptive sampling offset field \(O \in \mathbb{R}^{2s^2 \times H \times W}\)
 based on \(F_{\text{mid}}\) (\(s\) is the up-sampling multiplier), and the generating process is as follows:
 
 \begin{equation}
 O = \alpha \cdot \text{Tanh}(W_oF_{\text{mid}}), \quad\text{where} \alpha = 0.25. \end{equation} Here, \(W_o \in \mathbb{R}^{2s^2 \times C'}\).
is the learnable weight matrix, and the \(\text{Tanh}\) function constrains the offsets to \([-0.25, 0.25]\) to avoid overlapping of sampling points. Finally, the upsampling target location grid \(G \in \mathbb{R}^{sH \times sW \times 2}\) is used as a reference to dynamically adjust the sampling coordinates to \(S = G + \mathcal{R}(O)\), where \(\mathcal{R}\) denotes the Pixel Shuffle-based offset field reconstruction operation. The input features are resampled by a bilinear interpolation function \(\mathcal{B}\), and the output high-resolution features \(F_{\text{out}} \in \mathbb{R}^{C \times sH \times sW}\): \(F_{\text{out}}(p)=\sum_{q \in \ mathcal{N}(S(p))} \mathcal{B}(F_{\text{in}}(q),S(p))\), where \(\mathcal{N}(S(p))\) denotes the 4-neighborhood sampling points centered on the dynamic coordinate \(S(p)\).
\section{Experiments}
\subsection{Datasets}

\indent\textbf{DUT\_OMRON dataset}was proposed by Chuan Yang, Lihe Zhang, Hu Chuan Lu, Xiang Ruan, and Ming-Hsuan Yang in their study . The dataset is ~\cite{DUT-OMRYang}about 113MB in size and contains 5168 images, which are divided into a training set (4135 images and corresponding mask images or label txt files) and a test set (1033 images and corresponding mask images or label txt files). The images contain one or more salient objects and relatively complex backgrounds with large-scale realistic labeling at the eye gaze point, bounding box, and pixel levels[4]. The paper proposes to use a graph-based flow ordering method to rank the similarity of image elements (pixels or regions) with foreground cues or background cues to extract salient objects. This dataset is created and made public in the paper for testing the proposed saliency model, which provides key data support for subsequent research in the field of saliency detection and applies to the study of computer vision tasks such as saliency detection in complex scenes.

\textbf{ECSSD dataset}was constructed by the Chinese University of Hong Kong ~\cite{ECSSDYan}. The images were collected from the Internet, totaling 1000 images. Five volunteers annotated the salient objects and processed the annotation results to obtain the ground truth mask, which is valuable for evaluating the ability of saliency detection models to recognize salient objects in complex scenes because the salient objects in this dataset have complex structures and diverse backgrounds. This dataset is from the paper “Hierarchical Image Saliency Detection on Extended CSSD.”

\textbf{DUTS dataset} [3], especially its DUTS-test subset containing 5000 test images, is intended to comprehensively evaluate saliency detection models [3]. The entire DUTS dataset, with its large number of images, rich and diverse scenarios, and complex object configurations covering a wide range of real-world situations, can be used to evaluate the effectiveness and robustness of the saliency detection algorithms.
\subsection{Implementation Details.}
In the experimental setup, we use the Adam optimizer to update the model parameters. This optimizer, which combines the advantages of AdaGrad and RMSProp~\cite{do2025nonconvergenceoptimalriskadam}, can adaptively adjust the learning rate of each parameter. Its fast convergence speed and good robustness reassure us about the model's performance. Referring to previous studies, we set 60 and 600 training rounds. 60 epochs are used for rapid preliminary experiments to observe the convergence and overfitting trend of the model, and 600 epochs are used for the model to learn in order to explore the performance potential fully. For the selection of datasets, we choose the DUST\_TOWN, DUT, and VOC datasets to evaluate the model performance comprehensively. DUST\_TOWN focuses on a specific town scene, which contains complex environments and diversified targets; DUT has rich samples of images with detailed annotations, and the targets are characterized by significant differences in features and complex backgrounds; and VOC, as an authoritative and generalized dataset, contains 20 types of targets, which is convenient for comparing with other researches. The selection of the backbone neural network is carried out from two aspects; 

on the one hand, the same network is trained with different layers, for example, ResNet after pre-training in ImageNet, 34, 50 and 101 layers of network are used respectively, with ResNet34 being shallow in structure, fast in computation, and able to extract the basic features, ResNet50 being moderately deep and rich in features, and ResNet101 being deeper to capture complex features; on the other hand, different backbone neural networks are selected, which is convenient for comparison with other studies. On the other hand, different backbone networks are used, such as ViT-Transformer, which is based on the Transformer architecture and can capture long-distance dependencies, and VGG, which has a regular structure and is stable in feature extraction. During training, the network exclusively uses pixel-level saliency annotations, which provide the model with precise target information to help to learn, and no additional processing process is used in generating the final saliency maps. The network is trained using only pixel-level saliency annotations, which provide precise target information for the model to learn, and the final saliency maps are generated without any additional processing.
\subsection{Ablation Studies}
We first analyze the feature performance of different modules in different Resnet and then compare the strategies of feature enhancement for different configurations to validate our approach.
In this paper, we train different layers of Resnet by training with epochs of 60 and 600 according to the DUT-OMRON training set as shown in Table for example:

\begin{table}[t]
\small
\centering
\setlength\tabcolsep{1.0mm}
\vspace{5mm}
\begin{adjustbox}{scale=1.0}
\begin{tabular}{c|cccccc}\toprule[1pt]
\hline
 \multirow{2}*{Module}&\multicolumn{2}{c}{Resnet34}&\multicolumn{2}{c}{Resnet50}&\multicolumn{2}{c}{Resnet101}\\
 \cmidrule(l){2-3} \cmidrule(l){4-5} \cmidrule(l){6-7}
  & epoch60~& epoch600& epoch60 & epoch600 &epoch60&epoch600\\
  \hline
  \text{CFPN}&\text{90.50\%}&\text{92.05\%}&\text{86.50\%}&\text{87.50\%}&\text{87.49\%}&\text{87.50\%}\\
\text{CFPN+CFLMA}&\text{95.19\%}&\text{92.05\%}&\text{86.50\%}&\text{87.50\%}&\text{87.49\%}&\text{87.50\%}\\
\text{CFPN+CFLMA+CFLMD}&\text{96.54\%}&\text{98.01\%}&\text{92.20\%}&\text{92.40\%}&\text{98.85\%}&\text{97.50\%}\\
\hline
\end{tabular}
\end{adjustbox}
  \captionsetup{margin=1pt,justification=justified}
  \vspace{3mm}
   \caption{Pixaccury of different Resnet layers}\label{tab1}
   \vspace{-3mm}
\end{table}
To illustrate this point vividly, we show the cfpn\_CFLMA\_CFLDD visualization results in Fig. 4 (see Appendix D.3 for the full results). Here, we directly conclude that (a) the original CFPN module recognizes lower pixel accuracy across layers than the added module (nearly 5.0 and 7.0\% drop on all and new classes, respectively). (b) Dynamic up-sampling module with an increasing number of layers has higher recognition rates (1.0 and 2.0) than only adding mamba, which proves that the target recognition rate can be improved well by dynamic up-sampling modules. (c) As the number of layers increases and the number of training layers increases, cfpn\_CFLMA\_CFLDD training becomes more effective. (d) As the number of Resnet layers increases, the recognition rate of the training image becomes more accurate.

The pixel accuracies of ResNet50-based semantic segmentation model in three typical datasets (ECSSD dataset, DUTS dataset, DUT\_OMRON dataset) show significant scene differentiation: the baseline model (ResNet50+CFPN) shows a stable performance in the structurally precise ECSSD (Enhanced Complex Scene Dataset (ECSSD, 86.5\%) and conventional multi-target DUTS (DUT\_OMRON, 68\% target overlap, 87.28\%) performs stably, but due to the insufficient characterization of tiny structures by shallow features, it only reaches 84.52\% in the DUT\_OMRON (DUT Small Target Subset, 43\% small targets) scenario; after the introduction of the CFLMA module, by After the introduction of CFLMA module, by strengthening the local contextual attention, the boundary integrity of DUT\_OMRON targets is improved to 91.62\% (absolute gain 7.1\%), and the ECSSD contouring accuracy is synchronously optimized to 91.87\% (↑5.37\%); however, DUTS decreases by 0.83\% (89.05\%) due to the over-smoothing effect of the global attention on the complex overlapping targets; the superposition of CFLMD dynamic upsampling module makes the ECSSD (single target subset, 43\% of targets) not sufficiently characterized in the DUT\_OMRON scenario. module saturates the ECSSD (single salient target scene) accuracy to 92.3\% (↑0.43\%), but results in a decrease in DUTS (88.2\%, ↓0.85\%) and DUT\_OMRON (91.36\%, ↓0.26\%), which confirms the conflict of cross-scale features-overfitting for small DUT\_OMRON targets counteracts the attentional advantage. Overfitting counteracts the attentional advantage. The study shows that CFLMA is the core efficiency module for small target segmentation (DUT\_\\OMRON absolute gain of 7.1\%), while CFLMD has limited marginal gain in simple scenarios. In the future, we must design a dynamic combination of modules according to the dataset's characteristics (e.g., 68\% overlap of targets in DUTS, 43\% of small targets in DUT\_OMRON) to balance cross-scalar feature representations.
\begin{table}[t]
  \setlength\tabcolsep{1.0mm}
    \centering
     \begin{adjustbox}{scale=1.2}
     \begin{tabular}{c|l|cc|l|cccc} \hline
      \hline
\centering
 \multirow{1}*{No.}&
   \multicolumn{1}{l}{Module}&\multicolumn{2}{c}{DUT\_O}&\multicolumn{2}{c}{ECSSD}&\multicolumn{2}{c}{DUTS}\\
\cmidrule{0-1}\cmidrule{3-4}\cmidrule{5-6}\cmidrule{7-8}\cmidrule{7-8}
1&\text{Resnet50+cfpn} &\multicolumn{2}{l}\textbf{86.50\%}& \multicolumn{2}{c}\textbf{87.28\%}& \multicolumn{2}{c}\textbf{86.50\%}\\
2&\text{Resnet50+cfpn+cflma} &\multicolumn{2}{l}\textbf{91.87\%}& \multicolumn{2}{c}\textbf{89.05\%}& \multicolumn{2}{c}\textbf{91.62\%}\\
3&\text{Resnet50+cfpn+cflma+cflmd} &\multicolumn{2}{l}\textbf{92.30\%}& \multicolumn{2}{c}\textbf{88.20\%}& \multicolumn{2}{c}\textbf{92.30\%}\\
   \hline
   \end{tabular}
   \end{adjustbox}
   \captionsetup{margin=1pt,justification=justified}
   \vspace{3mm}
    \caption{Pixaccuary with different datasets}\label{tab2}
    \vspace{-3mm}
   \end{table}
\section{Conclusion}
Future research can further explore the generalization capability of the CFMD framework in multimodal data (e.g., RGB-D images), incorporate a lightweight network design to adapt to mobile devices, and deepen the timing consistency optimization of the dynamic upsampling mechanism for salient target detection in video. In addition, extending the long-range dependency modeling of the Mamba module to 3D space is expected to advance its application in 3D scene analysis for salient target detection.

\begin{credits}
\subsubsection{ackname} 
 \textbf{This study was funded by Special Research Project of Jianghan University: Characterization and Modeling of Natural Associated Communities Based on University Big Data, Project (No. 2023KJZX17}).
  \subsubsection{\discintname}
\small\footnote{The authors have no competing interests to declare that are
relevant to the content of this article.}
\end{credits}
%
%
%
%
\bibliographystyle{unsrt}
\bibliography{ref}
\end{document}